\documentclass[a4paper]{article}

\newif\ifpdf
\ifx\pdfoutput\undefined
\pdffalse
\else
\pdfoutput=1
\pdftrue
\fi

\usepackage{amssymb}
\ifpdf
\usepackage{color}
\usepackage{garamond}

\usepackage[pdftex]{graphicx}
\usepackage{hyperref}
\else

\usepackage[dvips]{graphicx}
\fi
\usepackage{theorem}
\usepackage[ruled,vlined]{algorithm2e}

\theoremheaderfont{\scshape}
\newtheorem{definition}{Definition}[section]

\newtheorem{theorem}[definition]{Theorem}
\newtheorem{lemma}[definition]{Lemma}
\newtheorem{corollary}[definition]{Corollary}
\newtheorem{example}{Example}
\newtheorem{problem}{Problem}

\def\endproof{\penalty-250%
\hbox to 6pt{\hfill}\hfill\llap{%
\vbox{\hrule\hbox{\vrule height6pt\hskip6pt\vrule}\hrule}}\medskip}

\SetKwFunction{Proj}{Project}
\SetKwFunction{ProjAC}{AC-Project}
\SetKwFunction{Ext}{Extend}
\SetKwFunction{ExtAC}{AC-Extend}
\SetKwFunction{GAC}{GAC}
\SetKwFunction{DAC}{DAC}
\SetKwFor{Procedure}{Procedure}{}{}%

\renewcommand{\FuncSty}[1]{\textsf{#1}}
\renewcommand{\ArgSty}[1]{\textrm{#1}}

\def\setN{{\mathbb{N}}}
\def\proj{{\!\downarrow\!}}

\title{Arc consistency for soft constraints}

\author{Martin Cooper\\
\textsf{cooper@irit.fr}\\
IRIT, France \and
Thomas Schiex\\
\textsf{tschiex@toulouse.inra.fr}\\
INRA, France}
\date{}
\begin{document}

\maketitle

\begin{abstract}
  The notion of arc consistency plays a central role in constraint
  satisfaction. It is known
  since~\cite{Schiex95a,Bistarelli95,Bistarelli97} that the notion of
  local consistency can be extended to constraint optimisation
  problems defined by soft constraint frameworks based on an
  idempotent cost combination operator. This excludes non idempotent
  operators such as $+$ which define problems which are very important
  in practical applications such as \textsc{Max-CSP}, where the aim is
  to minimize the number of violated constraints.
  
  In this paper, we show that using a weak additional axiom satisfied
  by most existing soft constraints proposals, it is possible to
  define a notion of \emph{soft arc consistency} that extends the
  classical notion of arc consistency and this even in the case of non
  idempotent cost combination operators. A polynomial time algorithm
  for enforcing this soft arc consistency exists and its space and
  time complexities are identical to that of enforcing arc consistency
  in CSPs when the cost combination operator is strictly monotonic
  (for example \textsc{Max-CSP}).
  
  A directional version of arc consistency, first introduced
  in~\cite{CooperFCSP} is potentially even stronger than the
  non-directional version, since it allows non local propagation of
  penalties. We demonstrate the utility of directional arc consistency
  by showing that it not only solves soft constraint problems on
  trees, but that it also implies a form of local optimality, which we
  call arc irreducibility.
\end{abstract}

\section*{Introduction}

Compared to other combinatorial optimisation frameworks, the CSP
framework is essentially characterised by the ubiquitous use of
so-called local consistency properties and enforcing algorithms among
which arc consistency is certainly preeminent.

The notion of local consistency can be  characterised by a set
of desirable properties: 
\begin{itemize}
\item local consistency is a relaxation of consistency, which means
  that for any consistent CSP there is an equivalent non empty locally
  consistent CSP.
\item this equivalent locally consistent CSP, which is unique, can be
  found in polynomial time by so-called enforcing or filtering
  algorithms.
\end{itemize}
Several papers have tried to extend the classical notion of arc
consistency to weighted constraint frameworks. In such frameworks, the
aim is to find an assignment that minimises combined violations. The
first work in this direction is probably~\cite{Rosenfeld76} which
defined arc consistency filtering for conjunctive (max-min) fuzzy CSP.

This extension was rather straightforward and one might be tempted to
think that this would be the case for other frameworks such as
\textsc{Max-CSP}, introduced in~\cite{Shapiro81,Freuder92}, where the
aim is to find an assignment which minimises the (weighted) number of
violated constraints.  This turned out not to be the case. Later works
tried to extend arc consistency in a systematic way using axiomatic
frameworks to characterise the properties of the operator used to
combine violations:
\begin{itemize}
\item the Semi-Ring CSP framework was introduced
  in~\cite{Bistarelli95,Bistarelli97}. In this work, the extension of
  arc consistency enforcing is induced by a generalisation of the
  fundamental relational operators such as projection, intersection
  and join.  The essential conclusion of this work is that extended
  arc consistency works as long as the operator used to combine
  violations is idempotent.  This includes the case of conjunctive
  fuzzy CSP (in which we try to minimise the violation of the most
  violated constraint) and also some other cases with partial orders.
  For \textsc{Max-CSP} and other related cases, the algorithm may not
  terminate and may also provide non equivalent CSPs.
  
\item the Valued CSP framework was introduced in~\cite{Schiex95a}.
  Here, the extension of the arc consistency property is essentially
  based on the notion of relaxation. The same conclusion as in the
  Semi-Ring CSP framework was reached for idempotent operators. For
  other frameworks such as \textsc{Max-CSP}, it was shown that the
  problem of checking the extended arc consistency property defines an
  \textsf{NP}-complete problem.
\end{itemize}
Parallel to these tentative extensions of arc consistency, other
research such as~\cite{Wallace94,Bennaceur98,Schiex98,Schiex99} tried
to provide improved lower bounds for \textsc{Max-CSP}. The idea of
extending arc consistency was abandoned in order to simply provide the
most important service, \textit{i.e.} the ability to detect that a CSP has no
solution whose cost is below a given threshold.

Globally, each of these proposals violates some of the desirable
properties of local consistency. In this paper we show that it is
possible, by the addition to the Valued CSP framework of a single
axiom, to define an extended arc consistency notion that has all the
desirable properties of classical arc consistency except for the
uniqueness of the arc consistency closure. It has also the pleasant
property that in the idempotent operator cases, it reduces to existing
working definitions and uniqueness is recovered.

It has been shown~\cite{Schiex00b} that a lower bound can easily be
built from any of the arc consistency closures and that this lower
bound generalises and improves upon existing lower
bounds~\cite{Wallace94,Bennaceur98,Schiex98,Schiex99}. In this paper,
we also consider a directional version of arc consistency that
improves lower bounds by propagating partial inconsistencies and not
only value deletions as~\cite{Schiex98,Schiex99}. In fact, we show
that directional arc consistency, first defined in~\cite{CooperFCSP},
defines a locally optimal lower bound.

\section{Notations and definitions}

A constraint satisfaction problem (CSP) is a triple $\langle X,D,C\rangle$. $X$
is a set of $n$ variables $X=\{1,\ldots,n\}$. Each variable $i \in X$ has a
domain of values $d_i \in D$ and can be assigned any value $a \in d_i$,
also noted $(i,a)$. $d$ will denote the cardinality of the largest
domain of a CSP. $C$ is a set of constraints. Each constraint $c_P \in
C$ is defined over a set of variables $P \subseteq X$ (called the scope of
the constraint) by a subset of the Cartesian product $\prod_{i\in P} d_i$
which defines all consistent tuples of values.  The cardinality $|P|$
is the arity of the constraint $c_P$. $r$ will denote the largest
arity of a CSP. We assume, without loss of generality, that at most
one constraint is defined over a given set of variables. The set $C$
is partitioned into two sets $C = C^1 \cup C^+$ where $C^1$ contains all
unary constraints.  For simplification, the unary constraint on
variable $i$ will be denoted $c_i$, binary constraints being denoted
$c_{ij}$. $e=|C^+|$ will denote the number of non unary constraints in
a CSP. If $J\subseteq X$ is a set of variables, then $\ell(J)$ denotes the
set of all possible labellings for $J$ \textit{i.e.}, the Cartesian
product $\prod_{i\in J}d_i$ of the domains of the variables in $J$. The
projection of a tuple of values $t$ onto a set of variables $V \subseteq X$
is denoted by $t_{\proj V}$.  A tuple of values $t$ satisfies a
constraint $c_P$ if $t_{\proj P} \in c_P$.  Finally, a tuple of values
over $X$ is a solution iff it satisfies all the constraints in $C$.

\section{Valued CSP}

Valued CSP (or VCSP) were initially introduced in~\cite{Schiex95a}. A
valued CSP is obtained by associating a valuation with each
constraint. The set $E$ of all possible valuations is assumed to be
totally ordered and its maximum element is used to represent total
inconsistency.  When a tuple violates a set of constraints, its
valuation is computed by combining the valuations of all violated
constraints using an aggregation operator, denoted by $\oplus$. This
operator must satisfy a set of properties that are captured by a set
of axioms defining a so-called \emph{valuation structure}.

\begin{definition}
  A {\em valuation structure} is defined as a tuple $\langle E,\oplus,\succcurlyeq\rangle$ such
  that:
  \begin{itemize}
  \item $E$ is a set, whose elements are called valuations, which is
    totally ordered by $\succcurlyeq$, with a maximum element denoted
    by $\top$ and a minimum element denoted by $\bot$;
  \item $E$ is closed under a commutative, associative binary
    operation $\oplus$ that satisfies:
    \begin{itemize}
    \item {\em Identity}\/: $\forall \alpha \in E, \alpha \oplus \bot = \alpha$;
    \item {\em Monotonicity}\/: $\forall \alpha,\beta,\gamma  \in E, (\alpha \succcurlyeq \beta) \Rightarrow
      \bigl((\alpha \oplus \gamma) \succcurlyeq (\beta \oplus \gamma )\bigr)$;
    \item {\em Absorbing element}: $\forall \alpha \in E, (\alpha \oplus \top)=\top$.
    \end{itemize}
  \end{itemize}
\end{definition}

When $E$ is restricted to $[0,1]$, this structure of a totally ordered
commutative monoid with a monotonic operator is also known in
uncertain reasoning, as a triangular co-norm~\cite{DPnormes}.

It is now possible to define valued CSPs. Note that, for the sake of
generality, rather than considering that a valuation is associated
with each constraint, as in~\cite{Schiex95a}, we consider that a
valuation is associated with each tuple of each constraint. As
observed in~\cite{Bistarelli99}, the two approaches are essentially
equivalent.

\begin{definition}
  A valued CSP is a tuple $\langle X,D,C,S\rangle$ where $X$ is a set of $n$
  variables $X=\{1,\ldots,n\}$, each variable $i \in X$ has a domain of
  possible values $d_i \in D$.  $C = C^1 \cup C^+$ is a set of
  constraints and $S = \langle E,\oplus,\succcurlyeq\rangle$ is a valuation
  structure. Each constraint $c_P \in C$ is defined over a set of
  variables $P \subseteq X$ as a function $c_P: \prod_{i\in P} d_i \to E$.
\end{definition}

An assignment $t$ of values to some variables $J \subseteq X$ can be simply
evaluated by combining, for all assigned constraints $c_P$ (i.e., such
that $P \subseteq J$), the valuations of the projection of the tuple $t$ on $P$:
\begin{definition}
  In a VCSP $V=\langle X,D,C,S\rangle$, the valuation of an assignment
  $t$ to a set of variables $J \subseteq X$ is defined by:
  \[\mathcal{V}_{V}(t) = \mathop\bigoplus_{c_P \in C, P \subseteq J}[c(t_{\proj P})] \]
\end{definition}

The problem usually considered is to find a complete assignment with a
minimum valuation.  Globally, the semantics of a VCSP is defined by
the valuations $\mathcal{V}(t)$ of assignments $t$ to $X$.

The choice of axioms is quite natural and is usual in the field of
uncertain reasoning. The ordered set $E$ simply allows us to express
different degrees of constraint violation. The commutativity and
associativity guarantee that the valuation of an assignment is
independent of the order in which valuations are combined. The
monotonicity of $\oplus$ guarantees that assignment valuations cannot
decrease when constraint violations increase.  For a more detailed
analysis and justification of the VCSP axioms, we invite the reader to
consult~\cite{Schiex95a,Schiex99} which also emphasise the difference
between idempotent and strictly monotonic aggregation operators $\oplus$.
\begin{definition} An operator $\oplus$ is
  idempotent if $\,\forall \alpha \in E, (\alpha \oplus \alpha) = \alpha$. It is strictly
  monotonic if $\,\forall \alpha,\beta,\gamma \in E, (\alpha \succ \beta) \land (\gamma \neq \top )\Rightarrow (\alpha \oplus
  \gamma) \succ (\beta \oplus \gamma)$
\end{definition}
As shown in~\cite{Schiex95a}, these two properties are incompatible as
soon as $|E| > 2$. The only valuation structures with an idempotent
operator correspond to classical and possibilistic CSP~\cite{Schiex92}
(min-max dual to the conjunctive fuzzy CSP framework) which use $\oplus =
\max$ as the aggregation operator.  Other soft CSP frameworks such as
\textsc{Max-CSP}, lexicographic CSP or probabilistic CSP use a
strictly monotonic operator.

Arc consistency enforcing must yield an equivalent problem, the
so-called arc-consistency closure. Several notions of equivalence were
introduced in~\cite{Schiex95a,Schiex99} that enabled us to compare
pairs of VCSP with different valuations structure.  In this paper, the
notion of equivalence will only be used to compare pairs of VCSP with
the same valuation structure and can therefore be simplified and
strengthened. 
\begin{definition}
  Two VCSP $V=\langle X,D,C,S\rangle$ and $V'=\langle X,D,C',S\rangle$ are equivalent iff
  for all complete assignment $t$ to $X$, we have:
  \[\mathcal{V}_{V}(t) = \mathcal{V}_{V'}(t) \]
\end{definition}

\section{Fair valuation structures}

We start with an introductory example. In the remainder of the paper,
in order to illustrate the notions introduced on concrete examples, we
will consider binary weighted \textsc{Max-CSP}s which correspond to
valued CSPs using the strictly monotonic valuation structure $\langle\setN \cup
\{\infty\}, +, \geq\rangle$.  To describe such problems, we use an undirected graph
representation where vertices represent values. For all pairs of
variables $i,j \in X$ such that $c_{ij}\in C$, for all values $a\in d_i$,
$b\in d_j$ such that $c_{ij}(a,b) \neq \bot =0$, an edge connect the values
$(i,a)$ and $(j,b)$. The weight of this edge is set to $c_{ij}(a,b)$.
Unary constraints are represented by weights associated with vertices,
weights equal to $0$ being omitted.

\begin{figure}[bthp]
  \begin{center}
    \includegraphics{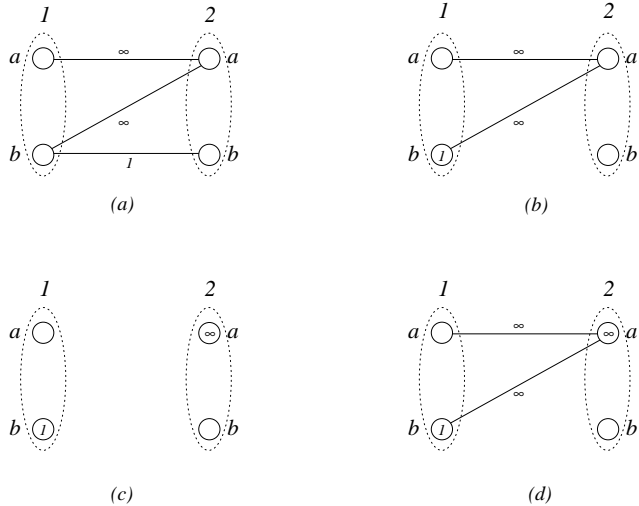}
    \end{center}
\caption{Four equivalent instances of \textsc{Max-CSP}\label{maxcsp1}}
\end{figure}

Let us consider the weighted \textsc{Max-CSP} in
figure~\ref{maxcsp1}(a). It has two variables numbered $1$ and $2$,
each with two values $a$ and $b$ together with a single constraint.
The constraint forbids pair $((1,b),(2,b))$ with cost $1$ and forbids
pairs $((1,a),(2,a))$ and $((1,b),(2,a))$ completely (with cost $\infty$).
The pair $((1,a),(2,b))$ is completely authorised and the
corresponding edge is therefore omitted.

If we assign the value $b$ to variable $1$, it is known for sure that
a cost of $1$ must be paid since all extensions of $(1,b)$ to variable
$2$ incur a cost of at least $1$. Projecting this minimum cost down
from c$_{12}$ would make this explicit and induce a unary constraint
on $1$ that forbids $(1,b)$ with cost $1$. However if we simply add
this constraint to the \textsc{Max-CSP}, as was proposed
in~\cite{Bistarelli95} for problems with an idempotent operator, the
resulting CSP is not equivalent. The complete assignment
$((1,b),(2,b))$ which initially had a cost of $1$ would now have a
cost of $2$.  In order to preserve equivalence, we must ``compensate''
for the induced unary constraint. This can be done by simply
subtracting $1$ from all the tuples that contain the value $(1,b)$.
The corresponding equivalent CSP is shown in figure~\ref{maxcsp1}(b):
the edge $((1,b),(2,b))$ of cost $1$ has disappeared (the associated
weight is now $0$) while the edge $((1,b),(2,a))$ is unaffected since
it has infinite weight.  We can repeat this process for variable $2$:
all extensions of value $(2,a)$ have infinite cost. Thus we can add a
unary constraint that completely forbids value $(2,a)$. In this
specific case, and because the valuation $\infty$ satisfies $\infty \oplus \infty =
\infty$, we can either compensate for this (Figure~\ref{maxcsp1}(c)) or
not (Figure~\ref{maxcsp1}(d)).  In both cases, an equivalent
\textsc{Max-CSP} is obtained. Between the problems in
Figure~\ref{maxcsp1}(c)) and~\ref{maxcsp1}(d), we prefer the problem
in Figure~\ref{maxcsp1}(d) because it makes information explicit both
at the domain and constraint level.

This type of projection mechanism underlies most of the lower bounds
defined for
\textsc{Max-CSP}~\cite{Wallace94,Bennaceur98,Schiex98,Schiex99}. To
our knowledge, the introduction of a ``compensation'' mechanism for
preserving equivalence was first introduced by~\cite{Koster99b} on
\textsc{Max-CSP}, independently of any notion of arc consistency. The
use of such mechanism for the definition and establishment of arc
consistency appeared in~\cite{Schiex00b} and in a related form
in~\cite{Horsch00} (for enforcing so-called probabilistic arc
consistency).

\begin{figure}[bthp]
  \begin{center}
    \includegraphics{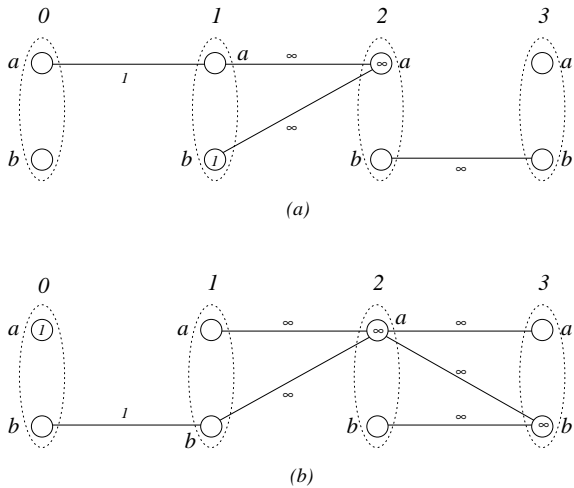}
    \end{center}
\caption{Two equivalent instances of \textsc{Max-CSP}\label{fig2}}
\end{figure}

Suppose now that the problem in Figure~\ref{maxcsp1} is part of an
instance of \textsc{Max-CSP} on four variables, as shown in
Figure~\ref{fig2}(a). As in crisp CSP, inconsistencies can propagate
from domains up to constraints. The cost of $\infty$ for $(2,a)$ can be
duplicated in the costs of the pairs $((2,a),(3,a))$ and
$((2,a),(3,b))$. Since $c_{23}(b,b)=\infty$, this in turn implies that
the assignment $(b,3)$ inevitably has a cost of $\infty$. No further
propagation of infinite costs can be performed.

A similar process can be applied to finite costs but one must take
care to compensate any cost change. The cost $1$ of $(1,b)$ can be
first shifted to the constraint $c_{01}$: the costs of the pairs
$((0,a),(1,b))$ and $((0,b),(1,b))$ become equal to $1$ and the cost
of the value $(1,b)$ is set to $0$.  Since now $c_{01}(a,a) =
c_{01}(a,b) = 1$, a cost of $1$ can be projected onto value $(0,a)$.
Figure~\ref{fig2}(b) shows the result of such propagations.  In the
case of finite costs, the process is obviously not terminated since
one could forever shift this cost back and forth between values
$(0,a)$ and $(1,b)$.

\subsection{A new axiom for VCSPs}

To formalise and generalise the ideas presented in the previous
section to other valuation structures, we have to be able to
compensate for the information added by projecting weights down onto
domains. This is made possible by the following additional axiom:

\begin{definition}\label{fairdef}
  In a valuation structure $S=\langle E,\oplus,\succcurlyeq\rangle$, if $\alpha,\beta \in E$,
  $\alpha \preccurlyeq \beta$ and there exists a valuation $\gamma \in E$ such
  that $\alpha \oplus \gamma = \beta$, then $\gamma$ is known as a difference of $\beta$ and
  $\alpha$.
  
  The valuation structure $S$ is \emph{fair} if for any pair of
  valuations $\alpha,\beta\in E$, with $\alpha \preccurlyeq \beta$, there exists a
  maximal difference of $\beta$ and $\alpha$. This unique maximal difference
  of $\beta$ and $\alpha$ is denoted by $\beta \ominus \alpha$.
\end{definition}

\begin{lemma}\label{lemma1}
  Let $S=\langle E,\oplus,\succcurlyeq\rangle$ be a fair valuation structure. Then
  $\forall u,v,w, \in E, w\preccurlyeq v$, we have $(v \ominus w) \preccurlyeq v$
  and $(u \oplus w) \oplus (v \ominus w) = (u \oplus v)$.
\end{lemma}
\begin{proof} By definition, $(v \ominus w) \oplus w = v$. From the monotonicity of
  $\oplus$, this proves that $(v \ominus w) \preccurlyeq v$ (this inequality
  becomes strict if $\oplus$ is strictly monotonic and $v\neq \top$). The
  second property follows from the commutativity and associativity of
  $\oplus$: we have $(u \oplus w) \oplus (v \ominus w) = u \oplus ((v \ominus w) \oplus w) = (u \oplus
  v)$.
\end{proof}

Most existing concrete soft constraint frameworks, including all those
with either an idempotent or strictly monotonic operator $\oplus$ are
fair.
\begin{example}
  If $\oplus$ is idempotent, then it can easily be shown that
  $\oplus=\max$~\cite{Schiex95a}. Classical CSPs can be defined as VCSPs
  over the valuation structure $S=\langle\{\bot,\top\},\max,\succcurlyeq\rangle$,
  where $\bot$ represents \textit{true} and $\bot$ \textit{false}. The
  operator $\oplus$ is also idempotent in possibilistic
  CSPs~\cite{Schiex92} which define a $\min$-$\max$ problem which is
  dual to the $\max$-$\min$ problem of conjunctive fuzzy
  CSPs~\cite{Rosenfeld76,CooperFCSP}. When $\oplus=\max$, we have
  $\ominus=\max$, since $\max(\max(\alpha,\beta),\alpha) = \beta$ whenever
  $\alpha\preccurlyeq \beta$. When $\alpha=\beta$, then any valuation $\gamma\prec \alpha$ is
  also a valid difference of $\beta$ and $\alpha$ but it is clearly not
  maximal.
\end{example}

\begin{example}
  In the strictly monotonic valuation structure
  $\langle\mathbb{N}\cup\{\infty\},+,\geq\rangle$, $\ominus$ is defined by $\beta \ominus \alpha = \beta -
  \alpha$ for finite valuations $\alpha,\beta\in \mathbb{N}, \alpha\leq \beta$ and $(\infty \ominus
  \alpha) = \infty$ for all $\alpha \in \mathbb{N}\cup\{\infty\}$. In the general case of
  any strictly monotonic operator $\oplus$, the difference operator may
  not exist in $E$, but it has been proved in~\cite{CooperFCSP} that
  the difference operator can always be constructed by embedding the
  valuation structure in a larger valuation structure derived from the
  set $E\times E$, where $(\beta,\alpha)$ represents the imaginary $\beta \ominus \alpha$.
  This can be compared with embedding $\mathbb{R}$ in $\mathbb{C}$ so
  as to allow us to take square roots of negative numbers. This
  construction is interesting for lexicographic CSPs~\cite{Schiex93c}
  for which differences are not always defined in the original
  valuation structure.  Another possible approach is to transform the
  lexicographic CSP into a VCSP on the valuation structure
  $\langle\mathbb{N}\cup\{\infty\},+,\geq\rangle$ using the simple transformation
  described in~\cite{Schiex95a}.
\end{example}

\subsection{Equivalence preserving transformations}

As it has been demonstrated in the examples of Figures~\ref{maxcsp1}
and~\ref{fig2}, it is possible to transform a \textsc{Max-CSP} into an
equivalent but different \textsc{Max-CSP} using local transformations
(involving only one non unary constraint). Such operations will be
called equivalence-preserving transformations:

\begin{definition}
  The \emph{subproblem} of a VCSP $V=\langle X,D,C,S\rangle$ on $J\subseteq X$ is the
  VCSP $V(J) = \langle J,D_J,C_J,S\rangle$, where $D_J=\{d_j~:~j\in J\}$ and $C_J
  = \{c_P \in C~:~P\subseteq J\}$.
\end{definition}

\begin{definition}
  For a VCSP $V$, an \emph{equivalence-preserving transformation} of
  $V$ on $J\subseteq X$ is an operation which transforms the subproblem of
  $V$ on $J$ into an equivalent VCSP. If $C_J = \{c_P \in C~:~P\subseteq J\}$
  contains only one non unary constraint, such an operation is called
  an \emph{equivalence-preserving arc transformation}.
\end{definition}

\begin{example} 
  The procedures \Proj\ and \Ext\, described in Algorithm~\ref{ext}
  are examples of equivalence-preserving transformations.
  
  \Proj\ transform a VCSP by shifting valuations from the tuples
  of a given non unary constraint $c_P$ to the value $(i,a)$, where $i\in
  P, a\in d_i$. In order to preserve equivalence, any increase at the
  unary level is compensated at the tuple level (line~1 of
  Algorithm~\ref{ext}).
  
  Conversely, \Ext\ shifts the valuation from value $(i,a)$ to the
  tuples of the constraint $c_P$ where $i\in P$. Again, the fairness of
  the valuation structure allows us to compensate for the possible
  increase of the tuple valuations by an operation at the unary level
  (line~2 of Algorithm~\ref{ext}).
\end{example}

\begin{algorithm}[htbp]
  \Procedure{\Proj{$c_P,i,a$}}{
    $\beta  \gets \min_{t \in \ell(P-\{i\})}(c_P(t,a))$\;
    $c_i(a) \gets c_i(a) \oplus \beta$\;
    \ForEach{($t\in \ell(P-\{i\})$)}
    {\lnl{comp}$c_P(t,a) \gets c_P(t,a) \ominus \beta$\;}
    }
  \BlankLine
  \BlankLine
  \Procedure{\Ext{$i,a,c_P$}}{
    \ForEach{($t \in \ell(P-\{i\})$)}{
      $c_P(t,a) \gets c_P(t,a) \oplus c_i(a)$\;
      }
    \lnl{abs}    $c_i(a) \gets c_i(a) \ominus c_i(a)$\;
    }
  \caption{Two basic equivalence-preserving arc-transformations\label{ext}}
\end{algorithm}

\begin{theorem}
  Given any fair VCSP $V=\langle X,D,C,S\rangle$, for any $c_P \in C^+$, $i\in P$,
  $a\in d_i$, the application of \Proj\ or \Ext\ on $V$ yields an
  equivalent VCSP.
\end{theorem}
\begin{proof}
  To demonstrate equivalence, it is sufficient to prove that the value
  of $c_P(t,a) \oplus c_i(a)$ is an invariant of \Proj{$c_P,i,a$} and
  \Ext{$i,a,c_P$}.  For any $t\in\ell(P-\{i\})$, let $\gamma$ be the initial
  value of $c_P(t,a)$ and $\delta$ the initial value of $c_i(a)$. After
  the execution of \Proj, we have $(c_P(t,a) \oplus c_i(a)) = (\gamma \ominus \beta)
  \oplus (\delta \oplus \beta) = \gamma \oplus \delta$.  After the execution of \Ext, we have
  $(c_P(t,a) \oplus c_i(a)) = (\gamma \oplus \delta) \oplus (\delta \ominus \delta) = \gamma \oplus \delta$. This
  proves the invariances.
\end{proof}

As the example of Figure~\ref{fig2} showed in the case of
\textsc{Max-CSP}, the iterated application of equivalence-preserving
transformations such as \Proj\ and \Ext\ does not necessarily lead to
a quiescent state. The two following sections show how a limited
application of carefully designed equivalence-preserving
transformations can guarantee that a quiescent state will always be
reached.

\section{Soft Arc consistency}

In classical CSPs, arc consistency enforcing always increases the
information available on each variable.  In the case of soft arc
consistency, application of arc transformations will be limited to
operations that either increase the information available at the
variable level or that increase information available at the
constraint level \emph{as long as they do not lower the information
  available at the variable level}.  In the next section, we try to
better characterise when this is possible.

\subsection{On the structure of valuation structures}

\begin{definition}
  In a valuation structure $\langle E,\oplus,\succcurlyeq\rangle$, an element $\alpha \in
  E$ is an absorbing element iff $\alpha \oplus \alpha = \alpha$.
\end{definition}
Absorbing elements can be duplicated without affecting valuations.
They can be propagated, in the same way as inconsistencies are in
crisp CSPs. Non-absorbing elements $\alpha$ can be shifted from one
constraint to another, but each addition of $\alpha$ must be compensated
by a subtraction elsewhere.

In a valuation structure $\langle E,\oplus,\succcurlyeq\rangle$, if $\oplus$ is
idempotent then all elements of $E$ are absorbing. If $\oplus$ is a
strictly monotonic operator then the only absorbing elements are $\bot$
and $\top$. Intermediate cases occur in the following examples:






\begin{example}
  Imagine the possible sentences for driving offences. Suppose that
  penalty points (up to a maximum of $12$) are awarded for minor
  offences, whereas serious offences are penalised by suspension of
  the offender's driving license for a period of $y$ years, for some
  positive integer $y$. A driver who accumulates $12$ penalty points
  receives an automatic one-year suspension of his/her license. The
  set of sentences can be modelled by a valuation structure
  $S=\langle E,\oplus,\succcurlyeq\rangle$ of the form:
  \[E=\{(p,0)~:~p\in\{0,\ldots,12\}\} \cup \{(0,y)~:~y\in\mathbb{N}^*\cup\{\infty\}\} \]

  \begin{eqnarray*}
    (p,y) \prec (p',y') & \Leftrightarrow & (y< y') \lor ((y = y'=0) \land (p< p'))\\
    &&\\
    (p,0)\oplus(p',0) & = & (\min(p+p',12),0)\\
    (p,y) \oplus (p',y') & = & (0,y+y')\hspace*{2cm}\textrm{if $(y+y'\neq 0)$}\\
  \end{eqnarray*}
  Note that $(12,0)\prec(0,1)$ even though they both give rise to a
  one-year license suspension. The penalty $(0,1)$ is deemed to be
  worse because it can be cumulated. For example $(0,1)\oplus(0,1) =
  (0,2)$, whereas $(12,0)\oplus(12,0) = (12,0)$. Apart from $\bot=(0,0)$ and
  $\top=(0,\infty)$, this valuation structure contains another absorbing
  valuation, namely $(12,0)$. This is a fair valuation structure since
  $\oplus$ has the following inverse operation $\ominus$:
  \begin{eqnarray*}
    (p,0)\ominus (p',0) & = & (p-p',0)\quad\textrm{if $p < 12$}\\
    (12,0) \ominus (p',0) & = & (12,0)\\
    (0,y) \ominus (p',y') &=& (0,y-y')
  \end{eqnarray*}
\end{example}

\begin{example}
  Another interesting case occurs if, for example, a company wants to
  minimise both financial loss $F$ and loss of human life $H$ if a
  fire should break out in its factory. Supposing that the company
  considers that no price can be put on human life, we must have
  \[(F,H) < (F',H') \Leftrightarrow (H<H') \lor (H=H' \land F<F')\]
If a financial loss of $F_{\max}$ represents bankruptcy, then
\[ (F,H)\oplus (F',H') = (\min\{F+F',F_{\max}\},H+H')\]
and $(F_{\max},0)$ is an absorbing element which is strictly less than
$\top$. Note that this valuation structure is not fair, since it is
impossible to define $\alpha = (0,1)\ominus (F_{\max},0)$ such that $\alpha\oplus
(F_{\max},0) = (0,1)$.
\end{example}

\begin{example}
  Consider a valuation structure $S=\langle\mathbb{N}\cup\{\infty,\top\},\oplus,\geq\rangle$
  composed of prison sentences. Sentences may be of $n$ years, life
  imprisonment (represented by $\infty$) or the death penalty
  (represented by $\top$). There is a rule that states that two life
  sentences lead automatically to a death sentence: in other words
  $(\infty\oplus\infty) = \top$.  Otherwise, sentences are cumulated in the obvious
  way: $\forall m, n \in\mathbb{N}, (m\oplus n = m+n)$; $\forall n\in \mathbb{N}, (\infty+n =
  \infty)$; $\forall \alpha\in E, (\top\oplus\alpha = \top)$. Although every pair $\beta,\alpha \in E,
  \alpha\leq \beta$ possesses a difference, this valuation structure is not
  fair since the set of differences of $\infty$ and $\infty$ is $\mathbb{N}$
  and hence no \emph{maximal} difference of $\infty$ and $\infty$ exists.
  However, $S$ can easily be rendered fair by replacing $\mathbb{N}$
  by $\{0,1,2,\ldots,150\}$, for example.
\end{example}

The following results show that all fair valuation structures are
composed of slices separated by absorbing values, each slice being
independent of the others.

\begin{lemma}\label{lemma2}
  Let $S=\langle E,\oplus,\succcurlyeq\rangle$ be a valuation structure. If $\alpha,\beta
  \in E$, $\alpha$ is an absorbing element and $\beta \preccurlyeq\alpha$ then
  $\alpha\oplus\beta = \alpha$. If $S$ is fair, then $\alpha \ominus \beta = \alpha$.
\end{lemma}
\begin{proof}
  Since $\beta\preccurlyeq \alpha$, it follows that $\alpha \preccurlyeq \alpha\oplus \beta
  \preccurlyeq \alpha\oplus\alpha =\alpha$, by monotonicity. Thus, $\alpha\oplus\beta = \alpha$.
  Furthermore, $\alpha\oplus\beta = \alpha$ shows that $\alpha$ is a difference of $\alpha$
  and $\beta$. It is the maximal difference since $\alpha\ominus \beta = (\alpha\ominus\beta)\oplus\bot
  \preccurlyeq (\alpha\ominus \beta)\oplus\beta = \alpha$, by monotonicity. Thus $\alpha\ominus\beta =\alpha$.
\end{proof}

\begin{lemma}\label{lemma3}
  Let $S=\langle E,\oplus,\succcurlyeq\rangle$ be a fair valuation structure. If
  $\alpha,\beta \in E$, $\alpha$ is an absorbing element and $\beta \succcurlyeq \alpha$
  then $\alpha\oplus\beta = \beta$ and $\beta \ominus \alpha = \beta$.
\end{lemma}
\begin{proof}
  Since $\alpha$ is absorbing, $\beta\oplus\alpha =(\beta\ominus \alpha)\oplus \alpha \oplus\alpha = (\beta\ominus \alpha)\oplus
  \alpha = \beta$. Furthermore, this shows that $\beta$ is a difference of $\beta$
  and $\alpha$.  It is the maximum difference since $\beta \ominus \alpha\preccurlyeq
  \beta$, by Lemma~\ref{lemma1}. Thus, $\beta \ominus \alpha = \beta$.
\end{proof}

\begin{theorem}[Slice Independence Theorem]\label{slicetheo}
  Let $S=\langle E,\oplus,\succcurlyeq\rangle$ be a fair valuation structure. Let
  $\beta,\gamma \in E, \beta\preccurlyeq \gamma$, and let $\alpha_0, \alpha_1\in E$ be absorbing
  valuations such that $\alpha_0\preccurlyeq \gamma\preccurlyeq\alpha_1$. Then
  $\alpha_0\preccurlyeq (\gamma \oplus\beta) \preccurlyeq \alpha_1$ and $\alpha_0
  \preccurlyeq (\gamma\ominus\beta) \preccurlyeq \alpha_1$.
\end{theorem}
\begin{proof}
  By monotonicity, $\beta\oplus\gamma \preccurlyeq \alpha_1 \oplus \gamma = \alpha_1$ by
  Lemma~\ref{lemma2}. By Lemma~\ref{lemma3}, $\gamma = \gamma \oplus \alpha_0 = (\gamma \ominus
  \beta) \oplus \beta \oplus \alpha_0 =(\gamma \ominus \beta) \oplus \alpha_0 \oplus \beta$.  Therefore, $((\gamma \ominus
  \beta) \oplus \alpha_0)$ is a difference of $\gamma$ and $\beta$. Since $\gamma\ominus\beta$ is a
  maximal difference, $\gamma\ominus\beta \succcurlyeq (\gamma\ominus\beta)\oplus\alpha_0\succcurlyeq
  \alpha_0$, by monotonicity. The remaining equalities follow from
  monotonicity.
\end{proof}
 
The following results will be useful for the proof of correctness of
the arc consistency enforcing algorithm given in the next section.
\begin{theorem}\label{maxabs}
  Let $S=\langle E,\oplus,\succcurlyeq\rangle$ be a fair valuation structure. For
  all $\alpha\in E$, $\alpha \ominus \alpha$ is the maximal absorbing valuation less than
  or equal to $\alpha$.
\end{theorem}
\begin{proof}
  Let $\beta=\alpha\ominus\alpha$. Now $\alpha\oplus(\beta\oplus\beta) = (\alpha\oplus\beta)\oplus\beta=\alpha\oplus\beta=\alpha$, which
  shows that $\beta\oplus\beta$ is a difference of $\alpha$ and $\alpha$.  By
  definition~\ref{fairdef}, $\beta$ is the maximal difference. Therefore,
  $\beta\succcurlyeq \beta\oplus\beta$. Since $\beta\preccurlyeq \beta\oplus\beta$ by
  monotonicity, we have $\beta=\beta\oplus\beta$ and hence  $\beta$ is
  absorbing. Maximality follows from Theorem~\ref{slicetheo}, since
  for all absorbing valuations $\alpha_0\preccurlyeq \alpha$, we have
  $\alpha_0\preccurlyeq (\alpha\ominus\alpha)$.
\end{proof}

\begin{lemma}\label{lemma4}
  Let $S=\langle E,\oplus,\succcurlyeq\rangle$ be a fair valuation structure. For
  all $\alpha,\beta\in E$, $((\alpha\oplus\beta)\ominus\alpha)\ominus\beta = (\alpha\oplus\beta)\ominus(\alpha\oplus\beta)$.
\end{lemma}
\begin{proof}
  Let $\gamma = (\alpha\oplus\beta)\ominus(\alpha\oplus\beta)$. By Theorem~\ref{maxabs}, $\gamma$ is
  absorbing. Now $\gamma\preccurlyeq (\alpha\oplus\beta)$, by Lemma~\ref{lemma1}. Let
  $\delta = ((\alpha\oplus\beta)\ominus\alpha)\ominus\beta$. Then the fact that $\delta \succcurlyeq \gamma$
  follows from two applications of the Slide Independence Theorem.
  But $\delta\oplus(\alpha\oplus\beta) = (\alpha\oplus\beta)$. Therefore $\delta$ is a difference of
  $(\alpha\oplus\beta)$ and $(\alpha\oplus\beta)$. $\gamma$ being the maximal difference, this
  shows that $\delta \preccurlyeq \gamma$ and $\delta=\gamma$.
\end{proof}

\begin{theorem}\label{theoabs}
  Let $S=\langle E,\oplus,\succcurlyeq\rangle$ be a fair valuation structure. For
  all $\alpha ,\beta \in E$, either $(\alpha\oplus\beta)\ominus \beta = \alpha$ or $(\alpha\oplus\beta)\ominus \beta =
  (\alpha\oplus\beta)\ominus(\alpha\oplus\beta)$ which is absorbing and strictly greater
  than $\alpha$.
\end{theorem}
\begin{proof}
  Let $\gamma = (\alpha\oplus\beta)\ominus(\alpha\oplus\beta)$. Now $(\alpha\oplus\beta)\ominus \beta =
  (((\alpha\oplus\beta)\ominus\beta)\ominus\alpha)\oplus\alpha = \gamma\oplus\alpha$, by Lemma~\ref{lemma4}. Since
  $\gamma$ is absorbing, $\gamma\oplus \alpha$ equals either $\alpha$ (if $\alpha \succcurlyeq
  \gamma$, Lemma~\ref{lemma3}) or $\gamma$ (if $\gamma \succcurlyeq \alpha$,
  Lemma~\ref{lemma2}). In this case, the fact that $\gamma$ is absorbing
  follows directly from Theorem~\ref{maxabs}.
\end{proof}

\begin{theorem}\label{finivs}
  Any VCSP $V=\langle X,D,C,S\rangle$ on a fair valuation structure $S=\langle
  E,\oplus,\succcurlyeq\rangle$ is equivalent to a VCSP on a valuation
  structure $S'$ with no more than $2ed^r+2$ absorbing valuations.
\end{theorem}
\begin{proof}
  For any $\beta \in E$ define
  \[\textit{Slice}(\beta) = \{\alpha\in E~:~\nexists \gamma \textrm{~absorbing in~}E \textrm{~s.t.~} (\alpha \prec \gamma \preccurlyeq \beta) \lor (\alpha \succ \gamma \succcurlyeq \beta) \}\]
  
  If $\beta$ is absorbing then $\textit{Slice}(\beta) = \{\beta\}$, otherwise
  $\textit{Slice}(\beta)$ is the set of valuations $\alpha$ for which there
  is no intermediate absorbing valuation $\gamma$ lying between $\alpha$ and
  $\beta$. Each $\textit{Slice}(\beta)$ contains at most two absorbing
  valuations, namely the maximum absorbing valuation less than or
  equal to $\beta$ (which is in fact $\beta \ominus \beta$, by Theorem~\ref{maxabs})
  and the minimum absorbing valuation greater than or equal to $\beta$
  (which may or may not exist).

  Let $E_0$ be the set of valuations taken on by the cost functions
  $c_P\in C$ in the VCSP and let

  \[E' = \{\bot,\top\} \bigcup_{\beta \in E_0} \textit{Slice}(\beta) \]
  
  Clearly $E'$ contains at most $2ed^r+2$ absorbing valuations. It
  is sufficient to show that $E'$ is closed under $\oplus$ and $\ominus$.
  
  Consider $\alpha,\beta \in E'$ such that $\alpha \preccurlyeq \beta$ and $\beta \in
  \textit{Slice}(\beta_o)$ for some $\beta_0 \in E_0$.
  
  Suppose that $(\alpha\oplus\beta) \not\in E'$. This implies that $\exists \gamma$
  absorbing in $E$ such that $\alpha\oplus\beta\succ \gamma\succcurlyeq \beta_0$. But,
  since $\beta\preccurlyeq\gamma$ by the definition of
  $\textit{Slice}(\beta_0)$, this contradicts Theorem~\ref{slicetheo}.
  Similarly, $\beta \ominus \alpha \not\in E'$ implies that $\exists\gamma$ absorbing in $E$
  such that $\beta\ominus\alpha\prec \gamma\preccurlyeq \beta_0$. But, since $\beta\succcurlyeq
  \gamma$ by the definition of $\textit{Slice}(\beta_0)$, this again
  contradicts Theorem~\ref{slicetheo}.
\end{proof}

By theorem~\ref{finivs}, we can now assume, without loss of
generality, that the number of absorbing valuations in the valuation
structure is finite.

\subsection{A Definition of Soft Arc Consistency}

Before giving an arc consistency enforcing algorithm which is valid
over any fair valuation structure, we require a formal definition of
arc consistency for fair VCSPs. We first consider the usual restriction
to binary VCSPs.

\begin{definition}\label{sac}
  A fair binary VCSP is \emph{arc consistent} if for all $i,j \in X$
  such that $c_{ij} \in C^+$, for all $a \in d_i$ we have:
\begin{enumerate}
\item\label{conda} $\forall b\in d_j, c_{ij}(a,b) = \bigl(c_i(a)\oplus c_{ij}(a,b)\oplus c_j(b)\bigr) \ominus \bigl(c_i(a)\oplus c_j(b)\bigr)$.
\item\label{condb} $c_i(a) = \min_{b\in d_j} (c_i(a) \oplus c_{ij}(a,b))$
\end{enumerate}
\end{definition}

Condition~\ref{conda} states that $c_{ij}(a,b)$ has been increased to
the maximal element in $E$ which does not increase the valuation
$\bigl(c_i(a)\oplus c_{ij}(a,b)\oplus c_j(b)\bigr)$ of $(a,b)$ on $\{i,j\}$.
If $\oplus$ is strictly monotonic or idempotent, then this is equivalent
to saying that absorbing valuations have been propagated from $c_i(a)$
to $c_{ij}(a,b)$.  Condition~\ref{condb} says that we
have propagated as much weight as possible from the constraint
$c_{ij}$ onto $c_i$. 

\begin{figure}
\begin{center}
\includegraphics[width=0.7\textwidth]{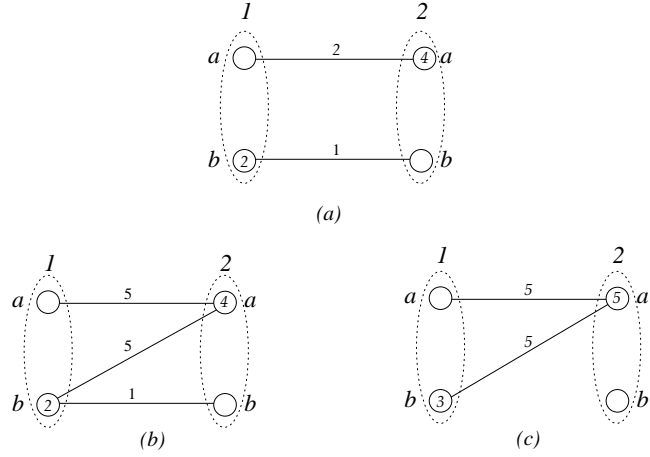}
\caption{(a) An example of a VCSP and how conditions~\ref{conda} and~\ref{condb} are enforced\label{figconda}}.
\end{center}
\end{figure}

To gain a better understanding of condition~\ref{conda} of
Definition~\ref{sac} in the most general case, consider a simple
valuation structure in which penalties lies in the range
$\{0,1,2,3,4,5\}$ and $\forall \alpha,\beta\in E, (\alpha\oplus\beta = \min(5,\alpha + \beta))$. $5$
is absorbing and verifies $5\ominus\alpha = 5$ for all $\alpha\preccurlyeq5$.
Figure~\ref{figconda}(a) shows a 2-variable VCSP over this valuation
structure.  Figure~\ref{figconda}(b) shows the result of enforcing
condition~\ref{conda} of Definition~\ref{sac}: $c_{12}(a,a)$ and
$c_{12}(b,a)$ can both be increased to $5$ without changing the
valuations of the solutions $(a,a)$ and $(b,a)$.
Figure~\ref{figconda}(c) shows the result of then enforcing
condition~\ref{condb}: penalties are projected down from constraints
to domains, as we have seen in the example of Figure~\ref{maxcsp1}.

Definition~\ref{sac} can be generalised to non binary VCSP. We call
this generalised arc consistency, to be consistent with the
terminology employed in the CSP literature~\cite{Mohr88}.

\begin{definition}\label{defgac}
  A fair VCSP is \emph{generalised arc consistent} if for all $c_P\in
  C^+$,  we have:
\begin{enumerate}
\item\label{gconda} $\forall t \in \ell(P), c_P(t) = (c_P(t) \oplus \beta) \ominus \beta$, where $\beta = \mathop\bigoplus_{j\in P} c_j(t_{\proj \{j\}})$.
\item\label{gcondb} $\forall i\in P, \forall a \in d_i, c_i(a) = \min_{t\in \ell(P-\{i\})} (c_i(a) \oplus c_P(t,a))$
\end{enumerate}
\end{definition}
Having given the necessary definitions, we can now define a
generalised arc consistency enforcing algorithm. 

\subsection{Enforcing generalised arc consistency in fair VCSPs}

Arc consistency is established by repeated calls to two subroutines
denoted by \ProjAC\ and \ExtAC\ (see Algorithms~\ref{algoproj}
and~\ref{algoext} respectively), called \emph{arc consistency
  operations}. The data structure $Q$ is a queue containing elements
$(t,P,\alpha)$, where $\alpha = c_P(t)$. The subroutine \ProjAC{$c_P,i,a$} is
a simple modification of the basic equivalence-preserving
transformation \Proj\ that memorises VCSP modifications in the queue
for further propagation. This simple modification obviously does not
alter the fact that it is an equivalence-preserving transformation.

\begin{algorithm}[htbp]
  \Procedure{\ProjAC{$c_P,i,a$}}{
    $\beta  \gets \min_{t \in \ell(P-\{i\})}(c_P(t,a))$\;
    \If{($c_i(a)\oplus \beta \succ c_i(a)$)}{
      $c_i(a) \gets c_i(a) \oplus \beta$\;
      Add $(a,\{i\},c_i(a))$ to $Q$\;
      \ForEach{($t\in \ell(P-\{i\})$)}
      {$c_P(t,a) \gets c_P(t,a) \ominus \beta$\;}
      }
    }
  \caption{Projection for Soft AC enforcing\label{algoproj}}
\end{algorithm}

\begin{lemma}\label{lemmaproj}
  For any fair VCSP $V=\langle X,D,C,S\rangle$, if $c_P\in C, i\in P, a\in d_i$ then
  the result of applying \ProjAC{$c_P,i,a$} to a fair VCSP $V$ is an
  equivalent VCSP in which
\[c_i(a) = \min_{t \in \ell(P-\{i\})} (c_i(a) \oplus c_P(t,a))\]
\end{lemma}
\begin{proof}
  This property follows from the fact that either $c_i(a)\oplus \beta$ is not
  strictly greater than $c_i(a)$ and in this case $c_i(a) = c_i(a)\oplus
  \min_{t \in \ell(P-\{i\})}(c_P(t,a)) = \min_{t \in \ell(P-\{i\})} (c_i(a)
  \oplus c_P(t,a))$ or else let $t_{\min}$ be the tuple such that
  $c_P(t_{\min},a) = \min_{t \in \ell(P-\{i\})}(c_P(t,a)) = \beta$ before
  execution of \ProjAC. Let $\delta$ be the original value of $c_i(a)$.
  After execution, $c_i(a) = \delta \oplus \beta = c_i(a) \oplus c_P(t_{\min},a)$.
  Therefore we have $c_i(a) \succcurlyeq \min_{t \in \ell(P-\{i\})}
  (c_i(a) \oplus c_P(t,a))$. The equality follows from monotonicity.
\end{proof}

The subroutine \ExtAC{$i,a,c_P$} is a modified version of the
equivalence-preserving transformation \Ext\ that propagates an
increase in the valuation $c_i(a)$ to all the valuations $c_P(t,a)$
for $t \in \ell(P-\{i\})$ when this can be done without any compensation
at the unary level.  It also memorises the new valuation $\gamma$ of each
tuple of $c_P$ in $Q$ for further propagation. In this case, $\gamma$ is
always an absorbing valuation (by Theorem~\ref{theoabs}).
 
\begin{algorithm}[htbp]
  \Procedure{\ExtAC{$i,a,c_P$}}{
    \ForEach{($t \in \ell(P-\{i\})$)}{
      $\beta \gets \mathop\bigoplus_{j\in P}c_j((t,a)_{\proj \{j\}})$\;
      $\gamma \gets (c_P(t,a)\oplus\beta) \ominus \beta$\;
      \If{($\gamma \succ c_P(t,a)$)}{
        $c_P(t,a) \gets \gamma$\;
        Add $((t,a),P,c_P(t,a))$ to $Q$\;
        }
      }
    }
  \caption{Extension for Soft AC enforcing\label{algoext}}
\end{algorithm}

\begin{lemma}\label{lemmaext}
  For any fair VCSP $V=\langle X,D,C,S\rangle$, if $c_P\in C, i\in P, a\in d_i$
  then the result of applying \ExtAC{$i,a,c_P$} to a fair VCSP $V$ is
  an equivalent VCSP in which
\[\forall t\in \ell(P-\{i\}), (c_P(t,a) = (c_P(t,a)\oplus \beta)\ominus \beta),\textrm{where~}\beta =\mathop\bigoplus_{j\in P} c_j((t,a)_{\proj \{j\}})\]
\end{lemma}
\begin{proof}
  To demonstrate equivalence, it is sufficient to prove that $\forall
  t\in\ell(P-\{i\})$, $\beta \oplus c_P(t,a)$ is an invariant of
  \ExtAC{$i,a,c_P$}, where $\beta =\mathop\bigoplus_{j\in P} c_j((t,a)_{\proj
    \{j\}})$. But this is certainly the case because, if $c_P(t,a) =
  \alpha$ before execution of \ExtAC{$i,a,c_P$}, then $\beta \oplus c_P(t,a) = \beta
  \oplus ((\alpha \oplus \beta)\ominus \beta) = \beta \oplus \alpha$ after $c_P(t,a)$ is updated by
  \ExtAC{$i,a,c_P$}.
  
  We know that $\alpha\preccurlyeq (\alpha\oplus \beta)\ominus\beta$, since
  $\alpha$ is clearly a difference of $\alpha\oplus\beta$ and $\beta$. If $\alpha\prec
  (\alpha\oplus\beta) \ominus \beta$, then $c_P(t,a)$ is assigned $(\alpha\oplus\beta) \ominus \beta$.
  Hence, after $c_P(t,a)$ is updated by \ExtAC{$i,a,c_P$}, $\beta \oplus
  c_P(t,a) = \beta \oplus \alpha$ and $(c_P(t,a) \oplus \beta) \ominus \beta = (\alpha \oplus \beta) \ominus \beta
  = c_P(t,a)$.
\end{proof}

We are now in a position to give an algorithm (Algorithm~\ref{refGAC})
for generalised arc consistency in fair VCSPs.

\begin{algorithm}
  \Procedure{\GAC{}}{
    \{ \emph{Initialisation phase} \}\;
    \ForEach{$i\in X$}
    {
      \ForEach{$a\in d_i$}
      {
        \ForEach{$c_P\in C^+$ s.t. $i\in P$}
        {
          \ExtAC{$i,a,c_P$}\;
          \ProjAC{$c_P,i,a$}\;
          }
        }
      }
    \{ \emph{Propagation phase} \}\;
    \While{$Q\neq\varnothing$}
    {
      Extract the first element $(t,P,\alpha)$ from $Q$\;
      \If{$c_P(t) = \alpha$}
      {
        \eIf{$P$ is a singleton $\{i\}$}
        {\lForEach{$c_P \in C^+$ s.t. $i\in P$}{\ExtAC{$i,t_{\proj \{i\}},c_P$}\;}}
        {\lForEach{$i\in P$} {\ProjAC{$c_P,i,t_{\proj\{i\}}$}\;}}
        }
      }
    }
  \caption{Generalised Arc Consistency enforcing for fair VCSPs\label{refGAC}}
\end{algorithm}

\begin{theorem}
When \GAC\ terminates, the resulting VCSP is generalised arc consistent.
\end{theorem}
\begin{proof}
  Consider $(c_P,i,a)$ such that $c_P \in C, i\in P, a\in d_i$. We know
  that \ProjAC{$c_P,i,a$} is called at least once during \GAC, since it
  is called in the initialisation phase. After the last call to
  \ProjAC{$c_P,i,a$}
\[c_i(a) = \min_{t \in \ell(P-\{i\})} (c_i(a)\oplus c_P(t,a))\]
by Lemma~\ref{lemmaproj}. This can only later become invalid by an
increase in some $c_P(t)$ by \ExtAC, which would necessarily be
accompanied by the addition of $(t,P,c_P(t))$ to $Q$ and would hence
entail another call of \ProjAC{$c_P,i,a$}. This contradiction
demonstrates that Condition~\ref{gcondb} in Definition~\ref{defgac} of
generalised arc consistency holds when \GAC\ terminates.

Consider $t\in\ell(P)$ where $c_P \in C^+$. We know that
\ExtAC{$i,t_{\proj \{i\}},c_P$} is called during the initialisation
phase for each $i \in P$. After the last such call of
\ExtAC{$i,t_{\proj\{i\}},c_P$} for any $i\in P$,
\[\forall t\in \ell(P), c_P(t) = (c_P(t) \oplus \beta) \ominus \beta, \textrm{where~}\beta =\mathop\bigoplus_{j\in P} c_j(t_{\proj \{j\}})\]
by Lemma~\ref{lemmaext}. This could only later become invalid by an
update of $c_P(t)$ or some $c_j(t_{\proj \{j\}})$ by
\ProjAC{$c_N,j,t_{\proj \{j\}}$} for some $c_N\in C$ such that $j\in P\cap
N$. But then a call of \ExtAC{$j,t_{\proj \{j\}},c_P$} would ensue. This
contradiction shows that Condition~\ref{gcondb} of
Definition~\ref{defgac} of generalised arc consistency also holds when
\GAC\ terminates.
\end{proof}

\begin{theorem}
\GAC\ has polynomial time complexity.
\end{theorem}
\begin{proof}
  Let $n_1$ be the number of elements $(t,P,\alpha)$ extracted from $Q$
  during \GAC\ such that $P$ is a singleton, and let $n_2$ be the
  number of elements $(t,P,\alpha)$ extracted from $Q$ during \GAC\ such
  that $|P|\geq 2$. By Theorem~\ref{finivs}, we can assume that the
  valuation structure contains at most $2ed^r+2$ absorbing valuations.
  
  Let $c_P\in C^+$ and $t\in \ell(P)$. By Theorem~\ref{theoabs}, \ExtAC\ 
  can only increase $c_P(t)$ to an absorbing valuation strictly
  greater than its previous valuation. \ProjAC\ can decrease $c_P(t)$
  but, by the Slice Independence Theorem, only from a non-absorbing
  valuation $\gamma$ to a valuation larger than or equal to $\delta = \gamma \ominus
  \gamma$, the maximal absorbing valuation less than or equal to $\gamma$.
  Thus the sequence of absorbing valuations taken on by $c_P(t)$
  during \GAC\ are strictly increasing. Thus for each of the $2ed^r+2$
  attainable absorbing valuations $\alpha$, $(t,P,\alpha)$ is added to $Q$ at
  most once.  Thus $n_2 \leq ed^r(2ed^r+2)$.
  
  Now $n_1$ cannot exceed the number of calls of \ProjAC\ during \GAC\ 
  since tuples $(t,P,\alpha)$ such that $P$ is a singleton are only added
  to $Q$ by \ProjAC. The number of calls of \ProjAC\ is clearly bounded
  above by $erd+n_2r$. Thus $n_1+n_2\leq erd+(r+1)n_2 = O(e^2d^{2r})$.
  Thus the total number of iterations of the while loop in \GAC\ is a
  polynomial function of $e$ and $d$. The results follows immediately.
\end{proof}

\begin{definition} 
  An arc consistent closure of a VCSP $V$ is a VCSP which is arc
  consistent and which can be obtained from $V$ by a finite sequence
  of applications of arc consistency operations \ExtAC\ and \ProjAC.
\end{definition}

\begin{figure}[bthp]
\begin{center}
  \includegraphics[scale=0.5]{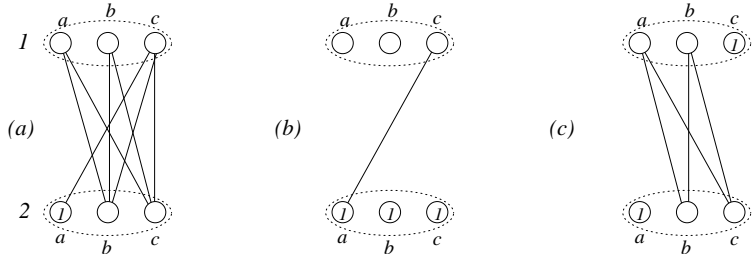}
\end{center}
\caption{A \textsc{Max-CSP} and two different equivalent arc consistent
  closures\label{uniq}}
\end{figure}

Note that confluence of arc consistency enforcing is lost and
therefore the arc consistent closure of a problem is not necessarily
unique as it is in classical CSPs. Figure~\ref{uniq}(a) shows a
2-variable VCSP on the valuation structure $\langle\setN\cup\{\infty\}, +,\geq\rangle$.
Each edge has a weight of $1$.  Figures~\ref{uniq}(b)
and~\ref{uniq}(c) show two different arc consistency closures of this
VCSP.

\subsection{Maximum arc consistency}

One of the practical use of arc consistency in VCSP is the computation
of lower bounds on the valuation of an optimal solution. Obviously,
given a VCSP $V$, the following valuation $f_{\min}(V)$ is always a
lower bound on the cost of an optimal solution:
\[  f_{\min}(V)= \mathop\bigoplus_{c_P \in C}[\min_{t\in \ell(P)} c_P(t)]\]
From this point of view, the closure in ~\ref{uniq}(b) is preferable
to the closure in Figure~\ref{uniq}(c) since it makes explicit the
fact that $1$ is a lower bound on the valuation of all solutions.

\begin{definition}
  A VCSP $V$ is said to be \emph{maximally arc consistent} if it is
  arc consistent and if the associated lower bound $f_{\min}(V)$ is
  maximum over all arc consistent closures of $V$.
\end{definition}

The problem of enforcing maximal arc consistency is certainly
practically important and some closely related problems are known to
be \textsf{NP}-hard~\cite{Schiex98a,Schiex95a}. Consider the following
problem:

\begin{problem}[Max-AC]
  Given a VCSP $V=(X,D,C,S)$ and a valuation $\kappa\in S$, does there
  exist an arc consistency closure $V'$ of $V$ such that $f_{\min} (V')
  \succcurlyeq \kappa$ ?
\end{problem}

\begin{theorem}
  The decision problem \textsc{Max-AC} is \textsf{NP}-complete.
\end{theorem}
\begin{proof}
  \textsc{Max-AC} is clearly in \textsf{NP}. It is therefore
  sufficient to give a polynomial reduction from \textsc{3-Sat} to
  \textsc{Max-AC}.
  
  Let $I_{\textsc{\scriptsize 3-Sat}}$ be an instance of \textsc{3-Sat},
  consisting of $n$ variables and $d$ clauses, each clause being the
  disjunction of exactly $3$ literals.
  
  We assume that each boolean variable $u$ and its negation $\lnot u$
  occur exactly the same number of times in $I_{\textsc{\scriptsize 3-Sat}}$. Note
  that if this is not initially the case, it can easily be achieved by
  adding the required number of tautological clauses of the form $(u
  \lor u \lor \lnot u)$ or $(u \lor \lnot u \lor \lnot u)$. We will now construct an instance
  $V$ of \textsc{Max-CSP} on $4d$ variables such that $V$ has an arc
  consistency closure $V'$ with $f_{\min}(V') \succcurlyeq
  \frac{5d}{2}$ iff $I_{\textsc{\scriptsize 3-Sat}}$ is satisfiable. Note that $d$
  is necessarily even by our assumption that each variable $u$ and its
  negation occur the same number of times.
  
\begin{figure}
  \begin{center}
    \includegraphics[width=0.9\textwidth]{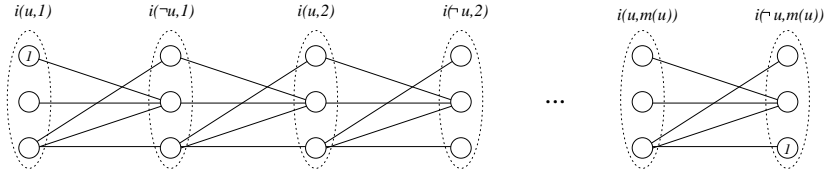}
    \caption{The gadget $G(u)$ representing the $m_u$ occurrences of $u$ and the $m_u$ occurrences of $\lnot u$ in $I_{\textsc{\scriptsize 3-Sat}}$\label{gadget-var}}
  \end{center}
\end{figure}

Suppose that the boolean variable $u$ (and its negation $\lnot u$) occur
in exactly $m(u)$ clauses in $I_{\textsc{\scriptsize 3-Sat}}$. Then we add the
gadget $G(u)$ shown in Figure~\ref{gadget-var} to $V$, containing the
$2m(u)$ variables $i(u,r)$, $i(\lnot u,r)$ ($r=1,\ldots,m(u)$), each with
domain-size $3$. Each edge joining value $a$ at variable $i$ and value
$b$ at variable $j$ represents a penalty of $1$, \textit{i.e.}
$c_{ij}(a,b) =1$. For each clause $c$ in $I_{\textsc{\scriptsize 3-Sat}}$, we add
the gadget $H(c)$ shown in Figure~\ref{gadget-cl}, involving a new
variable $i(c)$ connected to three existing variables. In the example
shown, $c\equiv (u\lor v\lor \lnot w)$ where $c$ contains the $p^{th}$ occurrence
of the boolean variable $u$ in $I_{\textsc{\scriptsize 3-Sat}}$, the $q^{th}$
occurrence of $v$ and the $r^{th}$ occurrence of $\lnot w$. In the gadget
$H(c)$ shown in Figure~\ref{gadget-cl}, $i(c)$ is connected to $i(\lnot
u,p)$, $i(\lnot v,q)$ and $i(w,r)$.

\begin{figure}[bhtp]
  \begin{center}
    \includegraphics[width=0.4\textwidth]{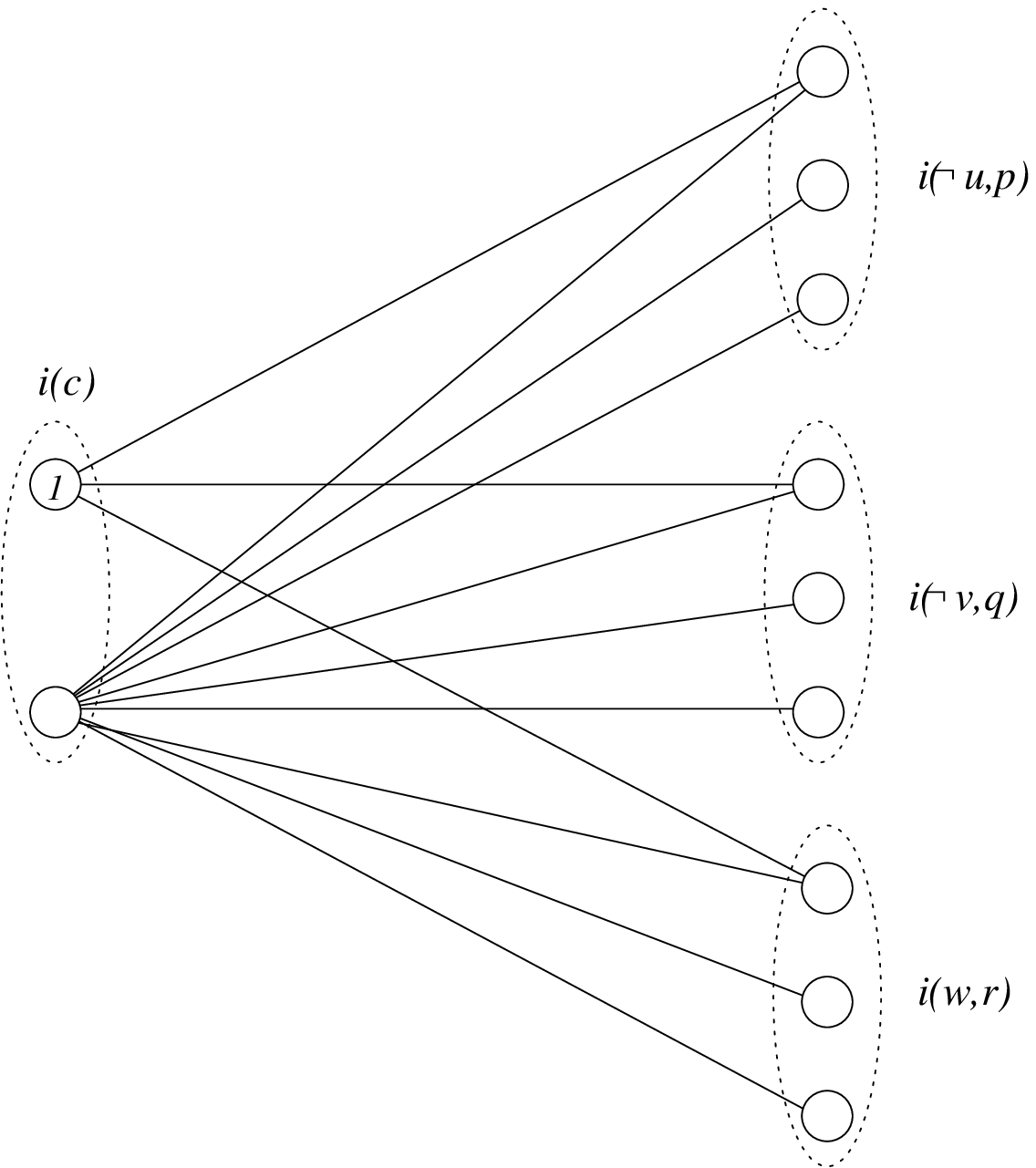}
    \caption{The gadget $H(c)$ for  $c\equiv (u\lor v\lor \lnot w)$.\label{gadget-cl}}
  \end{center}
\end{figure}

Let $M_i = \max_{a\in d_i}(c'_i(a))$, where $c'$ represents the
constraint functions in $V'$. Each variable $i(c)$ can contribute at
most $1$ to $f_{\min}(V')$, \textit{i.e.} $M_{i(c)} \leq 1$.  By
construction of $G(u)$, if any variable (\textit{e.g.} $i(u,r)$)
contributes $1$ to $f_{\min}(V')$, then its adjacent variables ($i(\lnot
u,r-1)$ and $i(\lnot u ,r)$) cannot contribute to $f_{\min}(V')$. This
means that the maximum value of $f_{\min}(V')$ is $d+\frac{3d}{2} =
\frac{5d}{2}$, since the total number of variables in the gadgets
$G(u)$ is $3d$. Indeed, $f_{\min}(V') = \frac{5d}{2}$ iff for all
clauses $c$, $M_{i(c)} = 1$ and for all boolean variables $u$,

    \begin{eqnarray}\label{eqn-gadget}
      (\forall r\in\{1,\ldots,m(u)\}, M_{i(u,r)} =1 \land M_{i(\lnot u,r)} = 0) & \lor\nonumber\\
      (\forall r\in\{1,\ldots,m(u)\}, M_{i(\lnot u,r)} =1 \land M_{i(u,r)} = 0)
    \end{eqnarray}

Consider the clause $c\equiv (u\lor v\lor \lnot w)$, whose gadget $H(c)$ is shown in Figure~\ref{gadget-cl}. By construction of $H(c)$, $M_{i(c)}=1$ implies that

\[(M_{i(\lnot u,p)} = 0) \lor (M_{i(\lnot v,q)} = 0) \lor (M_{i(w,r)} = 0) \]

which, in turn, implies from~(\ref{eqn-gadget}), that

\begin{equation}\label{eqn-gadget2}
  (M_{i(u,1)} = 1) \lor (M_{i(v,1)} = 1) \lor (M_{i(w,1)} = 0) 
\end{equation}

Suppose that $V$ has an arc consistency closure $V'$ with
$f_{\min}(V') \succcurlyeq \frac{5d}{2}$. For each variable $u$ in
$I_{\textsc{\scriptsize 3-Sat}}$, set $u = \textit{true}$ iff $M_{i(u,1)} = 1$.
For each clause $c$, for example $c\equiv (u\lor v\lor \lnot w)$, we know
from~(\ref{eqn-gadget2}) that either $u =\textit{true}$,
$v=\textit{true}$ or $w=\textit{false}$. Hence $I_{\textsc{\scriptsize 3-Sat}}$ is
satisfied.

Suppose that $\omega$ is a model of $I_{\textsc{\scriptsize 3-Sat}}$. In each
gadget $G(u)$ in $V$ and for each $r\in\{1,\ldots,m(u)\}$, if
$u=\textit{true}$ in $\omega$ then project penalties onto $i(u,r)$
from its constraints with the adjacent variables $i(\lnot u, r-1)$ and
$i(\lnot u,r)$; if $u=\textit{false}$ in $\omega$, then project penalties
onto $i(\lnot u,r)$ from its constraints with the adjacent variables
$i(u,r)$ and $i(u,r+1)$. Consider a gadget $H(c)$ in $V$, such as the
gadget illustrated in Figure~\ref{gadget-cl} for $c\equiv(u\lor v\lor\lnot w)$. If
$u=\textit{true}$ in $\omega$, then project penalties onto $i(c)$ from
its constraint with $i(\lnot u,p)$; if $u=\textit{false}$ in $\omega$,
then project penalties onto $i(\lnot u,p)$ from its constraint with
$i(c)$. Similarly, if $\lnot w =\textit{true}$ in $\omega$, then project
penalties onto $i(c)$ from its constraint with $i(w,r)$; if $\lnot w
=\textit{false}$ in $\omega$, then project penalties onto $i(w,r)$
from its constraint with $i(c)$.  Let $V'$ be the resulting arc
consistent VCSP. Since each clause is satisfied by $\omega$,
$M_{i(c)}=1$. Furthermore, if $u=\textit{true}$ in $\omega$, then
$M_{i(u,r)} =1$ for $r\in\{1,\ldots,m(u)\}$ and if $u=\textit{false}$ in
$\omega$, then $M_{i(\lnot u,r)} =1 $ for $r\in\{1,\ldots,m(u)\}$. Thus
$f_{\min}(V') = d+\frac{3d}{2} = \frac{5d}{2}$.
\end{proof}

\subsection{The case of  strictly monotonic VCSPs}

For strictly monotonic VCSPs, the previous algorithm can be improved
using an alternative equivalent definition of arc consistency based on
the notion of the underlying CSP.

\begin{definition}
  The \emph{underlying CSP} of a VCSP $V$ has the same variables and
  domains as $V$ together with, for each constraint $c_P\in C$, a crisp
  constraint $c'_P$ satisfying $\forall t\in \ell(P) (t\in c'_P \Leftrightarrow c_P(t) \prec \top)$
  (i.e., $t$ is not a totally forbidden labelling).
\end{definition} 

In strictly monotonic VCSPs, the only absorbing elements are $\top$ and
$\bot$. This allows us to give an equivalent but simpler definition of
generalised arc consistency:
\begin{theorem}\label{defsmgac} 
  If $\oplus$ is a strictly monotonic operator, then a VCSP is generalised
  arc consistent iff:
  \begin{enumerate}
  \item\label{smconda} its underlying CSP is generalised arc consistent
  \item\label{smcondb} $\forall c_P \in C^+, \forall i\in P, \forall a\in d_i$, if $c_i(a) \prec \top$ then $\exists t\in\ell(P-\{i\}) (c_P(t,a) = \bot)$.
  \end{enumerate}
\end{theorem}
\begin{proof}
  ($\Rightarrow$) Suppose that a strictly monotonic VCSP is generalised arc
  consistent but its underlying CSP is not. Then $\exists c_P \in C, i\in P,
  a \in d_i$ such that $(c_i(a) \prec \top) \land (\forall t\in \ell(P-\{i\}) (c_P(t,a)
  = \top \lor \exists j\in P-\{i\} (c_j(t_{\proj\{j\}}) = \top)))$. But, by
  Condition~\ref{gconda} of Definition~\ref{defgac},
  $(c_j(t_{\proj\{j\}}) = \top)$ implies $c_P(t,a) = \top$. Hence, $c_i(a)
  = \top$, by Condition~\ref{gcondb} of Definition~\ref{defgac}, which
  is a contradiction. Suppose on the other hand that the VCSP is
  generalised arc consistent but Condition~\ref{smcondb} of
  Theorem~\ref{defsmgac} is not satisfied. Then $\exists c_P \in C^+, i\in P,
  a\in d_i$ such that $c_i(a) \prec \top$ and $\forall t\in\ell(P-\{i\}), (c_P(t,a)
  \succ \bot)$. But by Condition~\ref{gcondb} of Definition~\ref{defgac},
  for some $t\in \ell(P-\{i\}), c_i(a) = c_i(a) \oplus c_P(t,a) \succ c_i(a)$ by
  strict monotonicity, which is impossible.\smallskip
  
  ($\Leftarrow$) Suppose that a strictly monotonic VCSP satisfies
  Condition~\ref{smconda} and~\ref{smcondb} of Theorem~\ref{defsmgac}.
  Condition~\ref{smcondb} clearly implies Condition~\ref{gcondb} of
  Definition~\ref{defgac}. Suppose that Condition~\ref{gconda} of
  Definition~\ref{defgac} is not satisfied. Then $\exists c_P\in C, t\in
  \ell(P)$ such that $c_P(t) \neq \delta = (c_p(t) \oplus \beta) \ominus \beta$ where $\beta =
  \mathop\bigoplus_{j\in P} c_j(t_{\proj \{j\}})$. By
  Theorem~\ref{theoabs}, $\delta$ is absorbing, which is only possible
  if $c_j(t_{\proj \{j\}})=\top$ for some $j\in P$. But the generalised
  arc consistency of the underlying CSP implies $c_P(t) = \top$, which
  provides the necessary contradiction.
\end{proof}

A possible way to enforce soft arc consistency on a strictly monotonic
VCSP is therefore to first enforce classical arc consistency on the
underlying CSP, assign a valuation of $\top$ to all deleted values in
the original VCSP and then enforce the second property by applying
\Proj{$c_P,i,a$} once for all $c_P \in C^+$, all $i \in P$ and all
$a\in d_i$ (see Algorithm~\ref{algoSACSM}).

\begin{algorithm}
  Enforce arc consistency in the underlying CSP\;
  \ForEach{$c_P\in C^+$}  {
    \ForEach{$i \in P$}{
      \lForEach{$a \in d_i$}{\Proj{$c_P,i,a$}\;}}
    }
  \caption{Enforcing soft arc consistency on strictly monotonic VCSPs\label{algoSACSM}}
\end{algorithm}

The only additional result needed to prove that this algorithm works
is the following one:
\begin{theorem}
  Let $V$ be a strictly monotonic VCSP whose underlying CSP is arc
  consistent. Then, $\forall c_P\in C^+, \forall i\in P, \forall a\in d_i$, the
  application of the equivalence-preserving transformation
  \Proj{$c_P,i,a$} yields a VCSP whose underlying CSP is unchanged
  (and therefore arc consistent).
\end{theorem}

\begin{proof}
  \Proj{$c_P,i,a$} cannot increase a valuation $c_i(a) \neq \top$ to $\top$
  since arc consistency on the underlying CSP would have deleted
  $(i,a)$ in this case and therefore we would have set $c_i(a) =\top$.
  It cannot decrease the valuation of any tuple such that $c_P(t,a) =
  \top$ since $\forall \beta\in E, \top \ominus \beta = \top$.
\end{proof}

\subsubsection{Improving space complexity}

If an optimal $O(ed^r)$ arc consistency enforcing algorithm such as
generalised AC7~\cite{Bessiere97} is used to enforce arc consistency
on the underlying CSP of a fair VCSP, Algorithm~\ref{algoSACSM}
establishes arc consistency in $O(ed^r)$ too.  However, the space
complexity of this algorithm is dominated by the space complexity of
the modified constraints which requires $O(ed^r)$ valuations. This is
extremely expensive, especially for constraints defined using a cost
function.  This space requirement can be reduced using a simple data
structure for representing modifications of costs induced by basic
equivalence-preserving transformations such as \Proj\ and \Ext.

Let us denote by $c^{or}_P \in C$ the original definition of a
constraint in a fair VCSP by any possible way: explicitly by a table
of valuations or implicitly by a cost function from $\ell(P) \to E$. For
each constraint $c_P$, for each variable $i\in P$, we use $2$ tables of
$d$ valuations noted $\Delta^+_{Pi}$ and $\Delta^-_{Pi}$, initialised to $\bot$.
Let $a\in d_i$:
\begin{itemize}
\item $\Delta^-_{Pi}[a]$ contains the combination of all the valuations
  that are projected from $c_P$ onto $(i,a)$;
\item $\Delta^+_{Pi}[a]$ contains the aggregation of all the valuations
  that are extended from $(i,a)$ to $c_P$.
\end{itemize}
At any time, the valuation of a tuple $t\in\ell(P)$ in the modified
constraint $c_P$ can simply be obtained by~:

\[ c_P(t) = c^{or}_P(t) \oplus (\mathop\bigoplus_{i\in P} \Delta^+_{Pi}[t_{\proj \{i\}}])  \ominus (\mathop\bigoplus_{i\in P} \Delta^-_{Pi}[t_{\proj \{i\}}])  \]

By definition of projection, $(\mathop\bigoplus_{i\in P} \Delta^-_{Pi}[t_{\proj
  \{i\}}]) \preccurlyeq (c^{or}_P(t) \oplus (\mathop\bigoplus_{i\in P}
\Delta^+_{Pi}[t_{\proj \{i\}}]))$ and the difference always exists in a
fair VCSP. The space complexity is now reduced to $O(edr)$ instead of
$O(ed^r)$ and our two basic equivalence-preserving transformations
\Proj\ and \Ext\ become space tractable even for large arity
constraints defined using cost functions.

\begin{algorithm}[htbp]
  \Procedure{\Proj{$c_{ij},i,a$}}{
    $\displaystyle\alpha  \gets \min_{t \in \ell(P-\{i\})}(c^{or}_P(t,a) \oplus (\mathop\bigoplus_{j\in P} \Delta^+_{Pj}[(t,a)_{\proj \{j\}}])  \ominus (\mathop\bigoplus_{j\in P} \Delta^-_{Pj}[(t,a)_{\proj \{j\}}]))$\;
    $c_i(a) \gets c_i(a) \oplus \alpha$\;
    $\Delta^-_{Pi}[a] \gets \Delta^-_{Pi}[a] \oplus \alpha$\;
    }
  \BlankLine
  \BlankLine
  \Procedure{\Ext{$i,a,c_P$}}{
    $\alpha  \gets c_i(a)$\;
    $\Delta^+_{Pi}[a] \gets \Delta^+_{Pi}[a] \oplus \alpha$\;
    $c_i(a) \gets c_i(a) \ominus \alpha$\;
    }
  \caption{Projection and Extension for Soft AC enforcing\label{Proj2}}
\end{algorithm}

Algorithm~\ref{Proj2} describes the procedures that implement these
transformations with these data structures.  The time complexity of
\Ext\ is reduced to $O(1)$ but since computing $c_P(t)$ requires
$O(r)$ $\oplus$ operations, \Proj\ is $O(rd^{r-1})$ instead of
$O(d^{r-1})$. This makes generalised arc consistency enforcing on
strictly monotonic VCSPs $O(red^r)$ in time. For binary constraints, we
recover the usual $O(ed^2)$ time and $O(ed)$ space complexities for
arc consistency enforcing.

\section{Directional arc consistency}

In CSPs, directional arc consistency is a weak version of arc
consistency. In VCSPs, the order imposed on variables by directional
arc consistency (first defined in~\cite{CooperFCSP} for strictly
monotonic operators) makes it possible to use the unlimited version of
\Ext (instead of \ExtAC) together with a terminating algorithm. For
this reason, soft directional arc consistency may be stronger than arc
consistency.

\begin{figure}[bhtp]
  \begin{center}
    \includegraphics[width=0.85\textwidth]{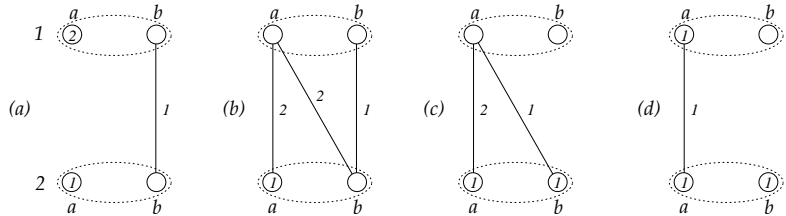}
  \end{center}
  \caption{Enforcing directional arc consistency\label{extp2}}
\end{figure}

Consider the Max-CSP in figure~\ref{extp2}(a).  It includes one binary
constraint that forbids pair $((1,b)(2,b))$ and two unary constraints
that forbid values $(1,a)$ and $(2,a)$. This VCSP is already arc
consistent and the corresponding lower bound $f_{\min}(V)$ is equal to
$0$.  However, we can apply \Ext\ on value $(1,a)$, getting the
equivalent VCSP~\ref{extp2}(b) which is not arc consistent.  We can
then apply \Proj\ on value $(2,b)$ and obtain the CSP~\ref{extp2}(c)
with a corresponding lower bound $f_{\min}(V) = 1$.

This improved lower bound has been obtained because we have decided to
pool all the unary valuations on one of the variables. This can be
done successively on all variables using any given variable order. In
the context of branch and bound (or other tree-based search), weights
can for example be propagated towards those variables which occur
earlier in the instantiation order.
\begin{definition}
  A binary VCSP is \emph{directional arc consistent} according to an
  order $<$ on variables if $\forall c_{ij} \in C^+$ such that $i < j$, $\forall
  a\in d_i$
  \[c_i(a) = \min_{b\in d_j} (c_i(a) \oplus c_{ij}(a,b) \oplus c_j(b))\]
\end{definition}

Provided that the VCSP is fair, directional arc consistency can be
established in polynomial time by the procedure \DAC\ in
Algorithm~\ref{algodac}, where \Proj\ and \Ext\ are as given in
Algorithm~\ref{Proj2}.

\begin{algorithm}
  \Procedure{\DAC{}}{
    \For{($i \gets (n-1)$ downto $1$)}
    {
      \ForEach{($j \in X$ s.t. $j> i$ and $c_{ij}\in C$)} {
        \{Start of propagation of $c_{ij}$\}\;
        \lForEach{($b\in d_j$)} {\Ext{$j,b,c_{ij}$}\;}
        \lForEach{($a\in d_i$)}{\Proj{$c_{ij},i,a$}\;}
        \{End of propagation of $c_{ij}$\}\;
        \lnl{assert}      \{Assert $A(i,j): \forall a\in d_i, c_i(a) =\min_{b\in d_j} (c_i(a)
        \oplus c_{ij}(a,b)\oplus c_j(b))$\}\;
        }
      }
    }
\caption{Enforcing directional arc consistency on fair binary VCSPs\label{algodac}}
\end{algorithm}

Since \DAC\ only applies equivalence-preserving transformations, it
yields an equivalent VCSP. The following lemma is needed to prove that
the VCSP obtained is directional arc consistent.

\begin{lemma}\label{lemmadac}
  If $\oplus$ is fair, then
  $$(\alpha = \alpha \oplus \beta \oplus \gamma) \land (\alpha' \succcurlyeq \alpha) \land (\gamma'\preccurlyeq \gamma) \Rightarrow (\alpha' = \alpha' \oplus \beta \oplus \gamma') $$
\end{lemma}
\begin{proof}
  Suppose that $(\alpha = \alpha \oplus \beta \oplus \gamma) \land (\alpha' \succcurlyeq \alpha) \land
    (\gamma'\preccurlyeq \gamma)$. Then $\alpha' \preccurlyeq \alpha'\oplus \beta \oplus \gamma'
    \preccurlyeq \alpha'\oplus \beta \oplus \gamma \preccurlyeq (\alpha' \ominus \alpha) \oplus (\alpha \oplus \beta \oplus
    \gamma) = (\alpha' \ominus \alpha) \oplus \alpha = \alpha'$. It follows that $\alpha' = \alpha' \oplus \beta
    \oplus \gamma'$.
\end{proof}

\begin{theorem}\label{theodac}
  If the binary VCSP is fair, then directional arc consistency can be
  established in $O(ed^2)$ time and $O(ed)$ space complexity.
\end{theorem}
\begin{proof}
  The assertion $A(i,j)$ is clearly true during execution of \DAC\ at
  line~1 of Algorithm~\ref{algodac} when constraint
  $c_{ij}$ has just been propagated.
  
  It suffices to show that $A(i,j)$ cannot be invalidated by later
  propagations of constraints $c_{i'j'}$ where $i'<j'$ and $(i'<i) \lor
  (i' = i \land j\neq j')$. Such operations may increase $c_i(a)$ and may
  decrease $c_j(b)$ but cannot modify $c_{ij}(a,b)$. From
  Lemma~\ref{lemmadac}, $c_i(a) = c_i(a) \oplus c_{ij}(a,b) \oplus c_j(b)$
  remains true and hence assertion $A(i,j)$ cannot be invalidated by
  later propagations and \DAC\ yields a directional arc consistent
  VCSP.
  
  As for time complexity, procedures \Ext\ and \Proj\ are called
  $O(ed)$ times and are both $O(d)$ for binary VCSP. \DAC\ is
  therefore $O(ed^2)$ in time. The $O(ed)$ space complexity can be
  attained using the implementations of \Proj\ and \Ext\ presented in
  Algorithm~\ref{Proj2}.
\end{proof}

When restricted to strictly monotonic VCSP, Theorem~\ref{theodac} can
be related to the result, proved in~\cite{CooperFCSP}, that
\emph{full directional arc consistency}, a stronger version of
directional arc consistency, can be established in $O(ed^2)$ time and
space complexity. A VCSP is full directional arc consistent if and
only if it is simultaneously arc consistent and directional arc
consistent.

\begin{theorem}
  Suppose that the constraint graph of a binary fair VCSP $V$ is a
  tree $T$ and that $V$ is directional arc consistent according to some
  topological ordering of the tree ($i$ is the father of $j$ in $T \Rightarrow
  (i < j)$). Then, for all $a\in d_1$, $c_1(a)$ is the optimal
  valuation over all solutions to $V$ in which variable $1$ is
  assigned value $a \in d_1$.
\end{theorem}

\begin{proof}
  Let $S$ be the set of all the sons of variable $1$ in $T$ and for
  each $j\in S$, let $T_j$ be the set of all variables in the subtree
  rooted in $j$.  By induction, we assume that $\forall j \in S,\forall b\in d_j$,
  the valuation $c_j(b)$ is the optimal valuation over all solutions
  to the subproblem on $T_j$. Let $t_{bj}$ be one corresponding
  optimal tuple over $T_j$.
  
  Let $a\in d_1$ and for each $j\in S$, let $b_{ja}$ be a value in $d_j$
  that minimises $c_1(a)\oplus c_{1j}(a,b_{ja}) \oplus c_j(b_{ja})$. Since $\forall
  i,j \in S, i\neq j$, $T_i \cap T_j = \varnothing$, we can build a tuple $t$ over
  $X$ by concatenation of each $t_{b_{ja}j}$ for all $j \in S$ and by
  assigning value $a$ to variable $1$.  The valuation of the tuple
  $t$, is $\mathcal{V}_V(t) = c_1(a) \oplus \mathop\bigoplus_{j\in S}\bigl(c_j(b_{ja}) \oplus c_{1j}(a,b_{ja})\bigr) =c_i(a)$ since the VCSP is
  directional arc consistent.
  
  Suppose there exists $t'$ such that $t_{\proj \{1\}} = a$ and
  $\mathcal{V}_V(t') \prec \mathcal{V}_V(t)$. Since the problem is
  tree-structured, the valuation of $t'$ can be written as 
  \[\begin{array}{rclr}
    \mathcal{V}_V(t')\!\!\! & =\!\!\! & c_1(a)  \oplus \mathop\bigoplus_{j\in S} \bigl(c_{1j}(a,t'_{\proj \{j\}}) \oplus \mathcal{V}_{V(T_j)}(t'_{\proj T_j})\bigr)\!\!\!\!\!\! & \textrm{(tree structure)}\\
                              & \succcurlyeq\!\!\! & c_1(a) \oplus  \mathop\bigoplus_{j\in S}\bigl(c_{1j}(a,t'_{\proj \{j\}}) \oplus c_j(t'_{\proj \{j\}})\bigr) & \textrm{(induction, monotonicity)}\\
                              & \succcurlyeq\!\!\! & c_1(a)  \oplus \mathop\bigoplus_{j\in S}\bigl(c_j(b_{ja}) \oplus c_{1j}(a,b_{ja})\bigr)& \textrm{(definition of $b_{ja}$)}\\
                              & =\!\!\! & c_1(a) = \mathcal{V}_V(t)\\
                            \end{array}\]
which shows that no such $t'$ exists.
\end{proof}

\section{Arc irreducibility}

We have seen that the cost of generalising arc consistency from crisp
to valued constraint satisfaction problems is the loss of uniqueness
of the arc consistent closure. As Figure~\ref{uniq} showed, two
different arc consistent closures may also induce different lower
bounds via $f_{\min}$.

If a VCSP $V$ is equivalent to another VCSP $V'$ which is better than
$V$ (according to some formally defined criterion $\mathcal{C}$) then
we say that $V$ is reducible for this criterion. Reducing a VCSP to an
equivalent irreducible problem is, in general, an \textsf{NP}-hard
problem. To see this, consider a CSP. If it is inconsistent
(\emph{i.e.}, has no solution), then its best equivalent problem, for
any reasonable criterion $\mathcal{C}$, is a CSP in which this fact is
made explicit, for example, by having $\forall a\in d_i, c_i(a) = \top$ for
some variable $i\in X$. Since testing consistency of a CSP is an
\textsf{NP}-complete problem, we can deduce that testing global
irreducibility is \textsf{NP}-hard.

Fortunately, irreducibility has local versions which are analogous to
local consistency. For example, a binary VCSP $V$ is arc-irreducible
if, for all pairs of variables $i,j\in X$, $V$ cannot be improved by
replacing $c_i,c_j,c_{ij}$ by an equivalent set of constraints
$c_i',c_j',c_{ij}'$. However, before giving a formal definition of
arc-irreducibility, we have to consider which criteria we could use to
compare equivalent VCSPs.

\begin{definition}
  A \emph{problem evaluation function} $f$ is a function which, for
  each VCSP $V$, assigns a value to $V$ in a totally ordered range.
  When comparing two equivalent VCSPs, $V_1$ and $V_2$, $V_1$ is
  considered as a better expression of the problem if $f(V_1) >
  f(V_2)$.
\end{definition}

\begin{example} The function $f_{\min}$ previously defined as
\[f_{\min}(V) = \mathop\bigoplus_{c_P\in C} ( \min_{t\in \ell(P)i} (c_P(t))) \]
is a problem evaluation function.
\end{example}

The main result presented in this section concerns $f_{\min}$, but the
definitions are valid for any problem evaluation function $f$.

\begin{definition}
  A binary VCSP $V$ is \emph{arc-irreducible} with respect to the
  problem evaluation function $f$ (or $(2,f)$-irreducible), if $\forall J
  \subseteq X, |J| =2$, for all VCSP $V'$ derived from $V$ by an
  equivalence-preserving transformation on $J$, $f(V) \geq f(V')$.
\end{definition}
Note that for certain choices of the problem evaluation function $f$,
an arc-irreducible VCSP is not necessarily arc-consistent. For
example, if $\forall i,j\in X, \exists a\in d_i, b\in d_j$ such that $c_i(a) =
c_j(b) = c_{ij}(a,b) = \bot$, then the VCSP is
$(2,f_{\min})$-irreducible but it is not necessarily arc-consistent.
Conversely, the VCSP in Figure~\ref{uniq}(c) is arc consistent, but
not $(2,f_{\min})$-irreducible.  The following theorem shows that
there is an important relationship between directional arc consistency
and $(2,f_{\min})$-irreducibility.

\begin{theorem}\label{dac2ir}
  A fair binary VCSP $V=\langle X,D,C,S\rangle$ which is directional arc consistent is
  $(2,f_{\min})$ -irreducible.
\end{theorem}
\begin{proof}
  Suppose that $V$ is directional arc consistent and let $i,j\in X$ be
  such that $i<j$. Then, by definition, $\forall a\in d_i,  \exists b\in d_j$ such that
  \[c_i(a) = \min_{b\in d_j} (c_i(a) \oplus c_{ij}(a,b) \oplus c_j(b))\]
  Thus the minimum valuation of a solution to the subproblem of $V$ on
  $\{i,j\}$ is
  \[\min_{a\in d_i}(c_i(a))\]
  Suppose that this minimum is attained for $a=u\in d_i$ and that $v\in
  d_j$ is such that
  \begin{equation}\label{eqdac}
    c_i(u) = (c_i(u) \oplus c_{ij}(u,v) \oplus c_j(v))
  \end{equation}
  Now consider a VCSP $V'$ obtained by an equivalence-preserving
  transformation of $V$ on $\{i,j\}$ which replaces $c_i, c_j,c_{ij}$
  by $c'_i, c'_j,c'_{ij}$. Then
\[f_{\min}(V') = \Bigl(f_{\min}(V) \ominus \bigl(\mathop\bigoplus_{P\subseteq \{i,j\}} [\min_{t\in \ell(P)} c_P(t)]\bigr)\Bigr) \oplus \bigl(\mathop\bigoplus_{P\subseteq \{i,j\}} [\min_{t\in \ell(P)} c'_P(t)] \bigr)\]
since $V$ and $V'$ only differ on $\{i,j\}$. Thus
\[\begin{array}{lcl}
f_{\min}(V')\!\!\!\! & \preccurlyeq\!\!\! & \bigl(f_{\min}(V) \ominus c_i(u)\bigr) \oplus \bigl( c'_i(u)\oplus c'_{ij}(u,v) \oplus c'_j(v)\bigr) \\
              & \preccurlyeq\!\!\! & \bigl(f_{\min}(V) \ominus  \bigl(c_i(u) \oplus c_{ij}(u,v) \oplus c_j(v)\bigr)\bigr) \oplus\bigl(c'_i(u) \oplus c'_{ij}(u,v) \oplus c'_j(v)\bigr)
\end{array}\]
by equation~\ref{eqdac}. Thus $f_{\min}(V') \preccurlyeq f_{\min}(V)$ due to the
equivalence of the subproblems of $V$ and $V'$ on $\{i,j\}$. Thus, $V$
is $(2,f_{\min})$-irreducible.
\end{proof}

This result must be considered with care. Given any VCSP $V$,
Theorem~~\ref{dac2ir}. states that any directional arc consistent closure
$V_{dac}$ of $V$ is always $(2,f_{\min})$-irreducible. However, an arc
consistent closure $V_{ac}$ may exist such that $f_{\min}(V_{ac}) \succ
f_{\min}(V_{dac})$.

\begin{corollary}
  Arc-irreducibility with respect to $f_{\min}$ can be established
  in $O(ed^2)$ time complexity and $O(ed)$ space complexity on fair
  binary VCSPs.
\end{corollary}
\begin{proof}
  This follows directly from Theorems~\ref{theodac} and~\ref{dac2ir}.
\end{proof}


\section*{Conclusion}

The concept of arc consistency plays an essential role in constraint
satisfaction as a problem simplification operation and as a
tree-pruning technique during search through the detection of local
inconsistencies among the uninstantiated variables. We have shown that
it is possible to generalise arc consistency to any instance of the
valued CSP framework provided the operator for aggregating penalties
has an inverse.

A polynomial-time algorithm for establishing soft arc consistency
exists. Its space and time complexity is identical to that of
establishing arc consistency in CSPs whenever the aggregation operator
of the VCSP is strictly monotonic, which is the case in
\textsc{Max}-CSP, for example. Contrarily to classical CSP arc
consistency, it does not define a unique arc consistency closure.
This algorithm nevertheless provides an efficient technique for
generating lower bounds on the value of a solution which can be used
during branch-and-bound search as in~\cite{Schiex98,Schiex99}. The
problem of finding the maximal lower bound is however
\textsf{NP}-hard.

We have also defined a directional version of soft arc consistency
which is potentially stronger since it allows non-local propagation of
penalties. Directional soft arc consistency implies a form of local
optimality in the expression of the VCSP, called arc
irreducibility. Furthermore, the complexity of establishing
directional arc consistency is identical to that of establishing arc
consistency in CSPs.


\end{document}